\documentclass{article}

\usepackage[final,nonatbib]{neurips_2025}

\usepackage[utf8]{inputenc} 
\usepackage[T1]{fontenc}    
\usepackage{hyperref}       
\usepackage{url}            
\usepackage{booktabs}       
\usepackage{amsfonts}       
\usepackage{nicefrac}       
\usepackage{microtype}      
\usepackage{xcolor}         

\usepackage{amsmath}
\usepackage{multirow}
\usepackage{booktabs}
\usepackage{adjustbox}
\usepackage{subcaption}
\usepackage{listings}
\usepackage{wrapfig}
\usepackage[table]{xcolor} 
\usepackage{algorithm}
\usepackage{algpseudocode}
\usepackage[numbers]{natbib}   

\algrenewcommand\algorithmicrequire{\textbf{Require:}}
\algrenewcommand\algorithmicensure{\textbf{Ensure:}}

\title{Decomposition of Small Transformer Models}
\workshoptitle{Mechanistic Interpretability Workshop at NeurIPS 2025
}

\author{%
  Casper L. Christensen\thanks{Casper L. Christensen was funded by the Pivotal research fellowship throughout this project.\\}\\\
  Independent\\ \url{casper.lutzhoft@gmail.com}\\
  \And
  Logan Riggs Smith\\
  Independent
}

\begin{document}

\maketitle

\begin{abstract}
Recent work in mechanistic interpretability has shown that decomposing models in parameter space may yield clean handles for analysis and intervention. Previous methods have demonstrated successful applications on a wide range of toy models, but the gap to "real models" has not yet been bridged.
In this work, we extend Stochastic Parameter Decomposition (SPD) to Transformer models, proposing an updated causal importance function suited for sequential data and a new loss function. 
We demonstrate that SPD can successfully decompose a toy induction-head model and recover the expected 2-step circuit. 
We also show that applying SPD to GPT-2-small can successfully locate subcomponents corresponding to interpretable concepts like "golf" and "basketball".
These results take the first step in the direction of extending SPD to modern models, and show that we can use the method to surface interpretable parameter-space mechanisms.
\end{abstract}

\section{Introduction}
Much of mechanistic interpretability work can be characterised as belonging to one of two waves \cite{sharkey2025lesswrong}. 
In the first, researchers attempted to understand models by dissecting and studying neurons individually, but this proved infeasible due to polysemanticity \cite{nguyen2016multifacetedfeaturevisualizationuncovering} and superposition \cite{elhage2022superposition}. 
The second wave shifted into \textit{activation space}, where sparse dictionary learning (SDL) \cite{sharkey2022lesswrong} uncovered thousands of highly-interpretable concepts \cite{cunningham2023sparseautoencodershighlyinterpretable}. 
However, SDL methods struggle with feature geometry and anomalies such as feature absorption and splitting \cite{chanin2025absorptionstudyingfeaturesplitting}, while offering no clear definition of what features actually are. 
Newer methods operate directly in parameter-space and provide a complementary line of work: Braun et al.~\cite{braun2025interpretabilityparameterspaceminimizing} decompose weights into ``mechanism-space`` vectors, and Bushnaq et al.~\cite{bushnaq2025stochasticparameterdecomposition} extend this with Stochastic Parameter Decomposition (SPD), which represents weights as sparsely-activating rank-1 matrices. In this work, we extend this method to Transformer models. Our contributions are: \textbf{a)} a causal importance function for sequential data, \textbf{b)} decomposing a toy model of induction heads, \textbf{c)} a loss function that enables faithful decomposition over few examples, and \textbf{d)} decomposing GPT2-small and locating interpretable subcomponents.

\section{Motivation} Mechanistic interpretability has mostly progressed through methods that operate on activations. Sparse dictionary learning, causal tracing, and circuit discovery let us ask, for a given input, “what fired and what did it do?” These approaches are powerful, but they do not yet give us a canonical way to factor the model itself into a small set of reusable mechanisms: features can split, and the geometry of the learned features is often left unexplained. Recent work on parameter-space decomposition argues that if gradient descent writes mechanisms directly into the weights, then it is natural to look for mechanisms in that same space.

\section{Parameter Decomposition} Attribution-based Parameter Decomposition (APD) makes this idea concrete: it flattens a trained network into parameter “components,” trains them to sum back to the original network (faithfulness), and then uses batch top-k to select the most highly attributed components, to make only a few components active on each example (minimality + simplicity). Stochastic Parameter Decomposition (SPD) keeps the same goal, but makes it more practical by working with rank-1 subcomponents and by learning a causal-importance function instead of hard-coding top-k. Concurrently, Chrisman et al. \cite{chrisman2025identifyingsparselyactivecircuits} propose Local Loss Landscape Decomposition (L3D), which also learns sparsely active parameter-space directions, but does so by reconstructing per-sample gradients of a divergence, rather than the weights themselves. Their sparsity is induced by a top-k reconstruction in gradient space, and their subnetworks are allowed to be higher-rank via Tucker decompositions. We directly extend the SPD line of work.

\section{Method}
SPD aims to provide interpretability by learning aa decomposed model that matches the original model's computations. Specifically, we want a decomposition that is \textbf{faithful} to the original model, \textbf{reconstructs} the original model's output for a given input $x$, , and is \textbf{minimal}.

SPD decomposes a model's weights (components) $W^l\in\mathbb{R}^{n \times m}$ into subcomponents $(W^l_1,...,W^l_C)$ where $C$ is a hyperparameter and each $W^l_c=\vec{U^l_c}\otimes\vec{V^l_c},~ U_c^{l},V_c^l\in\mathbb{R}^d$, such that they are rank 1. For any given input $x$, we can assemble a weight matrix $W'^l=\sum_{c=1}^C \alpha\cdot W_c^l\quad\alpha\in[0,1]$, which we can use instead of $W^l$. $\alpha$ then determines how "active" a subcomponent is on a given input. In practice, $\alpha \sim \mathcal{U}(g_c^l(x),1)$, where $g_c^l(x)$ is the causal importance of subcomponent $c$ in layer $l$ on $x$. We define $r^{s}$ as the mask of all $\alpha$-samples across all layers and $r^{l,(s)}$ as the mask of $\alpha$-samples for layer $l$. 

Faithfulness is enforced by requiring that all subcomponents sum to their original component. This means that the original model is the case where $\alpha=1$ everywhere. This ensures that we can always recover the real model from its decomposition, and restricts the decomposition's ability to learn some alternative factorisation of the original model's parameters. Let the model and its decomposition be respectively $f$ and $f'$, we then obtain a minimal decomposition by ensuring that $f(x)\approx f'(x)$ while assigning $g_c^l(x) > 0$ to as few subcomponents as possible.

This naturally leads to the losses from Bushnaq et al \cite{bushnaq2025stochasticparameterdecomposition}, and we train on a coefficient-weighted sum of them. See \ref{sec:step} for a description of a training step.
\begin{align}
    \mathcal{L}_{faithfulness} = \tfrac{1}{N}\sum_{l=1}^L\sum_{i,j}\left(W_{i,j}^{l}-\sum_{c=1}^{C}U_{i,c}^{l}V_{c,j}^{l}\right)^2,\quad\mathcal{L}_{minimality}=\sum_{l=1}^{L}\sum_{c=1}^{C}|g_c^l(x)|^p
    \label{eq1}
\end{align}
\begin{align} \mathcal{L}_{stochastic\_recon}=\frac{1}{S}\sum_{s=1}^SD_{KL}(f(x|W'(x,r^{(s)})),f(x|W))
    \label{eq2}
\end{align}
\begin{align}
\mathcal{L}_{stochastic\_recon\_layerwise}
= \frac{1}{L S}
\sum_{l=1}^L \sum_{s=1}^S
D_{KL}\!\left(f(x\mid W^1,\ldots,W^{l}(x,r^{l,(s)}),\ldots,W^L),\; f(x\mid W)\right)
    \label{eq3}
\end{align}
The stochastic sampling serves 2 purposes: it creates a mechanism by which subcomponents can still be turned on (and thus provide gradient signal) even if their causal importance became 0, and, because reconstruction should happen with as few subcomponents as possible, it sets the causal importance as the lower bound for how active a subcomponent can be without significantly changing the output. The layer-wise loss ensures that each decomposed weight must be substitutable in the original model without changing the output, further restricting the decomposition's ability to derive an equivalent model. In addition, we train on $\mathcal{L}_{recon}$, which is equivalent to $\mathcal{L}_{stochastic\_recon}$ with all mask-values set to exactly their causal importance (no sampling). The stochastic loss gives noisier signals, where the deterministic version tightens the lower bound and helps convergence.

\subsection{Causal Importances}
In Bushnaq et al \cite{bushnaq2025stochasticparameterdecomposition}, causal importance for a subcomponent is defined as how \textit{ablatable} it is. We follow this same convention, noting that ablations are not truly causal, because they are subject to interactions within the model \cite{covert2021explaining}. Within this definition, a causally important subcomponent should have high impact on the output and is therefore not ablatable. The choice of stochastic losses ensures that this is enforced, as non-important subcomponents may be arbitrarily turned on, whereas important ones are guaranteed to stay on. Causal importances are learned with an independent $\gamma$-function (an MLP) per subcomponent, which takes as input either the inner-activation $x \times \vec{U}$ (a scalar) or $x$. However, this is suboptimal for models operating on sequences. Consider the string "Sat by the river bank, the bank manager" and "manager" as the query-token. Independent $\gamma$-MLPs would be forced to assign the same causal importance score to the 2 instances of bank, even though the latter is likely to be more important for the output
\footnote{Note that this does not necessarily apply to the QK-circuit in Transformers with absolute positional encodings, as the representations will differ slightly. For RoPE encodings that are not applied until \textit{after} the projections, this will matter as the $\gamma$-MLPs will be positionally-unaware.}. This is especially true for the OV-circuit, where 2 value vectors with the same representation may be unequally attended to. Motivated by this, we apply attention with a minimal attention network (1 head, 1 layer, QKV-only) across the tokens prior to computing $g$. We use learned, relative positional encodings in the attention-network for expressivity and learned absolute positional encodings on the value-vectors, such that the downstream $\gamma$-MLP \textit{also} has positional awareness. This means that for subcomponents $(W_1^l,...,W_C^l)$ and positions ($1,...,n$), we decide causal importance $g^l_{c,n}$ via:

\begin{align}
g_{c,n}^l = \sigma_H\!\big( \gamma_c^l(\bar{x}_n) \big),\quad \text{where}~~
\bar{x_n}
= \left( \mathrm{softmax}\!\left( \frac{q_n K^\top + r_{n}}{\sqrt{d_k}} \right)V \right) \oplus x_n
\end{align}


where $V$ and $x_n$ have absolute, learned positional encodings, $r_n$ is the row of learned, relative positional biases, $\gamma_c^l$ are independent MLPs per subcomponent as in the previous literature, and $\sigma_H$ is the leaky-hard sigmoid function. While this scales worse than the original implementation, we implement this efficiently with flex-attention \cite{dong2024flexattentionprogrammingmodel}. See the appendix for a speed comparison.
\section{Experiments}
\subsection{Induction Heads}
\begin{figure}[h!]
    \centering
    \includegraphics[width=0.7\linewidth]{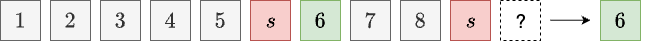}
    \caption{The type of sequence we train on for the induction-head task.}
    \label{fig:induction_task}
\end{figure}

Figure \ref{fig:induction_task} shows an example of the data generated for the task. Formally, we sample $n-2$ tokens from a vocabulary $V$, and we then insert 2 $s$-tokens; one randomly and one at the end of the sequence. The task is to predict the token following the first $s$-token (we call this token the $m$-token). The $s$-tokens are unique such that they serve as an indicator for induction. Crucially, this task is only solvable by first moving information from $s$ to $m$, and then $m$ to the final $s$, which forces the model to form an induction-head \cite{olsson2022context}. We train a 2 layer, 1-head-per-layer attention-only Transformer \cite{vaswani2023attentionneed} on sequences of this form, such that we can later decompose it. See Table \ref{tab:ih-hparams} for our hyperparameters. We consider loss only on the final token during model training and decomposition. We analyse the recovered circuit and its consistency with the original model.

\subsection{GPT2-Small}
One hope of methods like SPD is the ability to elicit latent knowledge. Decomposing over data points naturally creates a correspondence between the amount of data we decompose over and the proportion of the model we can "explain". However, in some cases, we may want to concretely investigate the model's behaviour on only a small amount of data points. To investigate this, we create a dataset of 2 sentences that GPT-2 correctly predicts the last token for: "Kobe Bryant plays the sport of basketball" and "Tiger Woods plays the sport of golf", and use these examples to decompose \textit{GPT2-small} \cite{Radford2019LanguageMA}. We analyse subcomponents unique to each example by suppressing their specific directions through orthogonal projection $W \;\leftarrow\; W \;-\; \sum_{k=1}^r 
\left( \hat{u}_k^\top W \, \hat{v}_k \right)
\hat{u}_k \, \hat{v}_k^\top,\quad\hat{u}_k = \frac{u_k}{\|u_k\|}, 
\quad
\hat{v}_k = \frac{v_k}{\|v_k\|}$. We also check this suppression on a 100-example synthetic fact set (\ref{sec:facts}) to show the edit is not purely single-example overfitting. 
When a decomposition covers only a small proportion of the original model, the unused components can instead become targets for the decomposed model to write new computation into. This can be more minimal if the decomposition develops a smaller circuit, but prevents us from interpreting the original model. We introduce randomised partial versions of the reconstruction losses that evaluate only a subset of decomposed layers at a time. This prevents the decomposed model from being able to rely on any of its subcomponents being consistently important.
\begin{align}
\mathcal{L}_{recon\_partial}
= D_{KL}\!\left(
f\!\big(x \mid W^{1},\ldots,W^{l \in \mathcal{M}}(x,g^{l}(x)),\ldots,W^{L}\big),\;
f(x \mid W)
\right),
\end{align}
\begin{align}
\mathcal{L}_{stochastic\_recon\_partial}
= \frac{1}{S} \sum_{s=1}^S 
D_{KL}\!\left(
f\!\big(x \mid W^{1},\ldots,W^{l \in \mathcal{M}^{(s)}}(x,r^{l,(s)}),\ldots,W^{L}\big),\;
f(x \mid W)
\right),
\end{align}
where $\mathcal{M}\subseteq \{1,...,L\}$. See \ref{sec:cheating} for further discussion on this approach.

\section{Results}
\subsection{Induction Heads}

\begin{table*}[ht]
\centering
\begin{subtable}[t]{0.4\textwidth}
    \centering
    \begin{adjustbox}{width=\linewidth}
    \begin{tabular}{l*{4}{c}}
        \toprule
        & $s_1$ & $m$ & $s_2$ & \textit{random} \\
        \midrule
        $Q_0$ & 0.000 & 1.000 & 0.000 & 0.001 \\
        $K_0$ & 1.000 & 0.050 & 0.000 & 0.183 \\
        $V_0$ & 1.000 & 0.000 & 0.000 & 0.000 \\
        \midrule
        $Q_1$ & 0.000 & 0.000 & 1.000 & 0.000 \\
        $K_1$ & 0.000 & 1.000 & 0.000 & 0.000 \\
        $V_1$ & 0.000 & 5.053 & 0.000 & 0.000 \\
        \bottomrule
    \end{tabular}
    \end{adjustbox}
    \caption{Average active subcomponents.}
    \label{tab:ih-active}
\end{subtable}
\hfill
\begin{subtable}[t]{0.2\textwidth}
    \centering
    \begin{adjustbox}{width=\linewidth}
    \begin{tabular}{lc}
        \toprule
        & Total unique \\
        \midrule
        $Q_0$ & 1 \\
        $K_0$ & 1 \\
        $V_0$ & 1 \\
        \midrule
        $Q_1$ & 1 \\
        $K_1$ & 1 \\
        $V_1$ & 11 \\
        \bottomrule
    \end{tabular}
    \end{adjustbox}
    \caption{Total unique subcomponents.}
    \label{tab:ih-unique}
\end{subtable}
\hfill
\begin{subtable}[t]{0.3\textwidth}
    \centering
    \begin{adjustbox}{width=\linewidth}
    \begin{tabular}{lc}
        \toprule
        Metric & Value \\
        \midrule
        $\mathcal{L}_{faithful}$ & $3e\!-\!9$ \\
        $\mathcal{L}_{recon}$ & $1e\!-\!4$  \\
        $\mathcal{L}_{recon\_layerwise}$ & $1e\!-\!4$  \\
        $\mathcal{D}_{KL}Attn$ (Layer 1) & 0.385 \\
        $\mathcal{D}_{KL}Attn$ (Layer 2) & 0.002 \\
        $\mathcal{D}_{KL}Attn$ (Mean) & 0.194 \\
        \bottomrule
    \end{tabular}
    \end{adjustbox}
    \caption{Evaluation metrics.}
    \label{tab:ih-metrics}
\end{subtable}
\caption{Results for the induction-head toy model decomposition}
\label{tab:ih-results}
\end{table*}

\begin{wrapfigure}{r}{0.5\linewidth} 
    \centering
    \includegraphics[width=\linewidth]{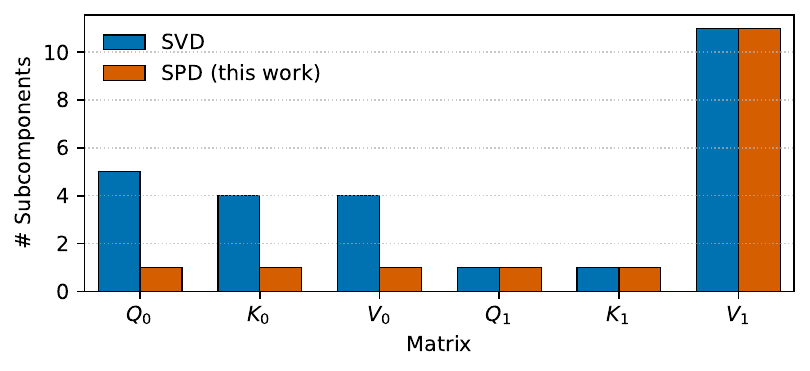}
    \caption{SPD vs. greedy SVD low-rank approximation (Algorithm \ref{alg:greedy-rank1}). Even on a very simple model, SPD finds a more minimal decomposition that matches the full model's output distribution.}
    \label{fig:svdspd}
\end{wrapfigure}
For the induction head toy model, only two positions matter per layer; $m$ must first attend to $s_1$ and 'understand' that it follows $s_1$, and then $s_2$ must attend to $m$ to obtain $m$'s identity. We see in Table \ref{tab:ih-results}a that subcomponents are assigned largely to the relevant positions, and Table \ref{tab:ih-results}b shows a very small amount of unique subcomponents. Table \ref{tab:ih-results}c shows almost-zero faithfulness, which indicates that if we consider all subcomponents causally important, we recover exactly the original model. $\mathcal{L}_{recon}$ and $\mathcal{L}_{recon\_layerwise}$ are equivalent to substituting all the weight matrices in the original; either all at once or layer-by-layer. This can be compared to ablating all the presumed-unimportant components of a model and seeing if it still works.

Investigating the concrete mechanisms, we learn that $K_0$'s objective is to align the representation of token $n$ with the positional encoding of $n+1$. When we do not add positional encodings prior to projecting $m$ through the Q-matrix, the attention-strength from $m$ to $s_1$ drops from $1.0$ to $0.0941$, even if $s_1$'s key gets to retain its positional information. $Q_1$ and $K_1$ implement the circuit in which $s_2$ attends to $m$, and we see considerably more active subcomponents in $V_1$ for $m$'s position. We believe this is because $m$'s identity among $128$ tokens must be unambiguously separable in the residual stream at the final position, and therefore higher rank than 1 is required to carry that information.
\subsection{GPT2-Small}
\begin{figure}[h!]
    \centering
    \begin{minipage}{0.48\linewidth}
        \centering
        \includegraphics[width=\linewidth]{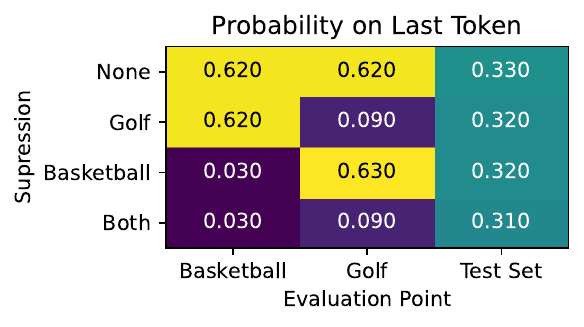}
        \caption{Ablating the slices associated with recovered facts has significantly higher effect on the specific data points.}
        \label{fig:ablation}
    \end{minipage}\hfill
    \begin{minipage}{0.48\linewidth}
        \centering
        \includegraphics[width=\linewidth]{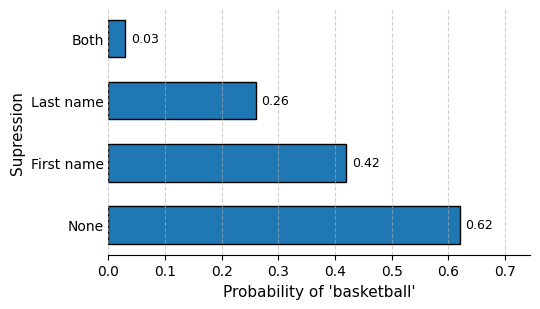}
        \caption{Both the first and last name for Kobe Bryant appear to carry basketball information and removing a rank 1 slice from both is most effective.}
        \label{fig:kobe_tokens}
    \end{minipage}
\end{figure}
For the 2 examples, we find 96 active subcomponents in total, which is a $99\%$ size reduction compared to the whole model \footnote{One caveat: Causal importance of subcomponents in the attention-mechanism. For instance, if a QK-circuit ends up with 0 active subcomponents, that circuit is still used, but attention is uniform (softmax over 0-vector). For this reason, we also do not compare sparsity or size with e.g pruning methods.}. We find 2 subcomponents that uniquely activate in the layer 0 MLP for tokens "obe" and "Bryant" and 1 uniquely for "Woods", that significantly reduce the probability of the sport being correctly predicted, if we suppress their directions in the original model. We see in figure \ref{fig:ablation} that these effects are isolated, which suggests we may have discovered where the network "stored" that fact. This extends previous work \cite{meng2023locatingeditingfactualassociations} which initially suggested facts may be stored closer towards the middle of the network, but where follow-up work \cite{hase2023doeslocalizationinformediting} came to the same conclusions we do, namely that you \textit{can} edit a fact somewhere the network does not store it. Our results suggest some facts originate as early as the very first MLP layer, and propagate through the network. Interestingly, we find that for the "Kobe Bryant" example, we must suppress a rank-1 direction on both the first and last name, to  maximally reduce the output probability. We believe this is because "Kobe" has a strong enough association with basketball by itself, whereas "Tiger" generally does not refer to a name. Importantly we suppress only the athlete $\rightarrow$ sport direction. If we reverse the direction and prompt with "The most famous athlete in golf is..", the model accurately answers "Tiger Woods". This is consistent with previous research \cite{meng2023locatingeditingfactualassociations} \cite{ berglund2024reversalcursellmstrained}, which finds that knowledge is stored asymmetrically in language models.
\section{Discussion}
While our results suggest that SPD can be extended to Transformer models, several limitations remain. First, our experiments are restricted to small-scale models, leaving open questions about scalability to larger architectures. Second, the causal importance parameterisation introduces additional computational overhead. Third, our evaluation is qualitative and task-specific, so further benchmarks are needed to establish generality across architectures and tasks. We also note the difficulty in establishing ground truth even for small, toy Transformers. While we obtain evidence that suggests our decompositions may be faithful, this is largely obtained through ablations.

Our results show that parameter decomposition can surface semantically specific mechanisms in Transformers. 
For induction heads, SPD recovers the expected two-step copy circuit. 
For GPT-2-small, ablating up to 2 rank-1 slices selectively reduces the probability of the targeted fact (e.g., ``golf’’ for Tiger Woods) while leaving unrelated examples largely unaffected. 
This indicates that SPD yields narrow, causally relevant directions rather than broad capacity that would degrade performance indiscriminately. 

These findings suggest that parameter-space objects can act as clean handles for mechanistic interventions. In future work we plan to extend this work to larger models, and to improve our definition of causal importance for computations with feature interactions.  

\newpage
\section*{Acknowledgements}
We thank Lucius Bushnaq, Dan Braun, and Lee Sharkey for frequent debate, and for open-sourcing the development of SPD and allowing us to contribute. Morgan Simpson, Tobias Häberli, Euan McLean, and Tilman Räuker were responsible for facilitating this research through the Pivotal Fellowship and also provided research management -- we are grateful to all of you. We also thank Rachael DeVries for extensive feedback on the manuscript. Finally, we thank Riya Tyagi, Kyle Reynoso, Frederik Hytting Jørgensen, and Jasmina Urdshals for giving feedback on earlier versions of this work and participating in brainstorming.
\bibliographystyle{plain} 
\bibliography{refs} 

\newpage
\appendix
\section{Appendix}
\subsection{Causal Importance Algorithm}
\begin{algorithm}[h]
\caption{Minimal Attention Causal Importance}
\label{alg:attn-doublepos}
\begin{algorithmic}[1]
\Require Input $x \in \mathbb{R}^{B\times s\times d}$
\Require $\Gamma^L$ = $(\gamma^l_1,...,\gamma^l_C)$ // A function that assigns $C$ causal importances per layer
\Require $\textit{Learnable}~\text{RelPosEnc}\in \mathbb{R}^{2\cdot s \times d}$
\Require $\textit{Learnable}~\text{AbsPosEnc}\in \mathbb{R}^{s \times d}$\Require $\textit{Learnable}~Q,K,V~\in \mathbb{R}^{d \times d}$
\Ensure Output $G^l \in \mathbb{R}^{B\times s\times C}$
\State $x_{pos} \gets x + \text{AbsPosEnc}(x)$
\State $q \gets Q(x)$;\quad $k \gets K(x)$;\quad $v \gets V(x_{pos})$
\State $r \gets \text{RelPosEnc}(s)$
\State $x_{combined} \gets \mathrm{softmax}(\tfrac{qk^\top + r}{\sqrt d})v$ \quad // Implemented with Flex Attention
\State $x_{combined} \gets \text{concat}(x_{combined}, x_{pos})$
\State $G^l \gets \Gamma^l(x_{combined})$
\State \Return $G^l$
\end{algorithmic}
\end{algorithm}

\begin{table}[h!]
    \centering
    \begin{tabular}{lcccc}
    \toprule
        & \multicolumn{4}{c}{Sequence Length} \\
        \cmidrule(lr){2-5}
        Method & 16 & 64 & 1024 & 10240* \\
    \midrule
        Scalar  & 12 & 12 & 9 & 11 \\
        Vector  & 13 & 13 & 10 & 11 \\
        Attention (no flex) & 10 & 10 & 2 & 1 \\
        Attention (flex)    & 10 & 10 & 6 & 6 \\
    \bottomrule
    \\
    \end{tabular}
    \\
    
    \caption{Iterations per second for different $\gamma$-MLP inputs on the induction-head decomposition task. *Batch size modified from 64 to 4 for sequence length 10240 due to memory limits.}
    \label{tab:placeholder}

\end{table}

\newpage
\subsection{Induction-Head Transformer}
\subsubsection{Hyperparameters}
\begin{table}[h!]
    \centering
    \begin{minipage}[t]{0.47\textwidth}
        \centering
        \begin{tabular}{lr}
            \textbf{Model hyperparameters} & \\
            \toprule
            Batch Size & 1024 \\
            Optimizer & AdamW \\
            Learning Rate & 0.001 \\
            Weight Decay & 0.01 \\
            Learning Rate Schedule & Constant \\
            Learning Rate Warmup & 1000 \\
            Steps & 100000 \\
            \midrule
            $D_{model}$ & 16 \\
            $Pos_{enc}$ & Shortformer \\
            Use Layer Norm & False \\
            Use FF & False \\
            \midrule
            Vocab Size & 128 \\
            Sequence Length & 64 \\
            \bottomrule\\
        \end{tabular}
        \caption{Hyperparameters for the Induction Head toy-model task}
        \label{tab:ih-hparams}
    \end{minipage}\hfill
    \begin{minipage}[t]{0.47\textwidth}
        \centering
        \begin{tabular}{lr}
            \textbf{Decomposition hyperparameters} & \\
            \toprule
            Batch Size & 1024 \\
            Optimizer & Adam \\
            Learning Rate & 0.001 \\
            Learning Rate Schedule & Cosine \\
            Learning Rate Warmup & 0 \\
            Steps & 100000 \\
            \midrule
            C & 100 \\
            $D_{gate}$ & 16 \\
            CI-function & Attention \\
            \midrule
            Stochastic Recon $\beta$ & 1 \\        
            Stochastic Layerwise Recon $\beta$ & 1 \\
            Recon $\beta$ & 0.5 \\
            Faithfulness $\beta$ & 1000 \\        
            Importance Minimality $\beta$ & 0.02 \\
            $P_{start}$ & 0.9 \\
            $P_{final}$ & 0.1 \\
            Anneal P? & True \\
            Last Token Only? & True \\
            \bottomrule\\
        \end{tabular}
        \caption{Hyperparameters for the Induction Head decomposition task}
        \label{tab:ih-decomp-hparams}
    \end{minipage}
\end{table}

\subsubsection{Loss Curve}
\begin{figure}[h!]
    \centering
    \includegraphics[width=1\linewidth]{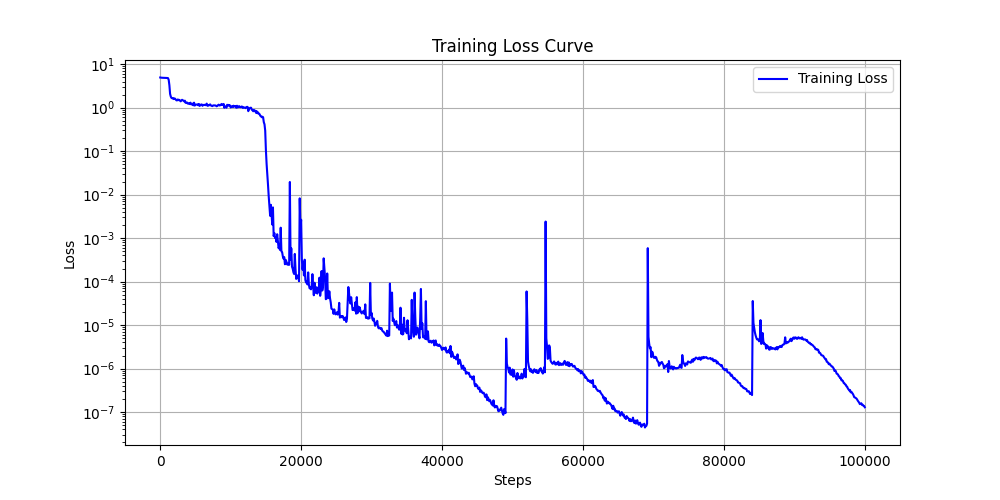}
    \caption{Loss curve for the Induction Head Transformer showing that expected phase changes are present.}
    \label{fig:placeholder}
\end{figure}
\subsubsection{SVD Baseline}
\begin{algorithm}[H]
\caption{Greedy Rank-1 SVD Pruning}
\label{alg:greedy-rank1}
\begin{algorithmic}[1]
\Require Weight matrices $\{W_i\}$ from target model $\mathcal{M}$
\Require Dataset $\mathcal{D} = \{x_b\}_{b=1}^N$
\Require Tolerance $\tau$
\Require Loss function $\mathcal{L}$~~~~~// $D_{KL}$ in our case.
\Ensure Reduced-rank weight matrices $\{\tilde{W}_i\}$

\State Decompose each $W_i$ with SVD: $W_i = U_i S_i V_i^\top$
\State $\text{current\_loss} \gets 0$
    \State $P \gets \mathcal{M}(\mathcal{D})$
\While{$\text{current\_loss} \le \tau$}
  \For{$W_i$ that still has rank $> 0$}: 
      \State Form $W_i'$ by removing its smallest remaining singular value/vector
      \State Define $\tilde{\mathcal{M}}$ as $\mathcal{M}$ with $W_i$ replaced by $W_i'$
      \State $Q \gets \tilde{\mathcal{M}}(D)$
      \State Evaluate $L_i = D_{KL}(P||Q)$
  \State Pick index $j$ with lowest $L_j$
  \If{$L_j \le \tau$}
    \State Update $W_j \gets W_j'$ in $\mathcal{M}$
    \State $\text{current\_loss} \gets L_j$
  \Else
    \State \textbf{break}
  \EndIf
\EndFor
\EndWhile

\State \Return pruned weight matrices $\{\tilde{W}_i\}$
\end{algorithmic}
\end{algorithm}

\newpage
\subsection{GPT-2}
\subsubsection{Hyperparameters}

\begin{figure}[h!]
 \centering
        \begin{tabular}{lr}
            \textbf{Decomposition hyperparameters} & \\
            \toprule
            Batch Size & 2 \\
            Optimizer & AdamW \\
            Learning Rate & 0.001 \\
            Weight Decay & 0.05 \\
            Learning Rate Schedule & Constant \\
            Learning Rate Warmup & 0.01 \\
            Steps & 20000 \\
            \midrule
            C & 768 \\
            $D_{gate}$ & 64 \\
            CI-function & vector \\
            \midrule
            Stochastic Recon $\beta$ & 0.1 \\    
            Partial Stochastic Recon $\beta$ & 10 \\      
            Stochastic Layerwise Recon $\beta$ & 0.1 \\
            Recon $\beta$ & 0.1 \\
            Partial Recon $\beta$ & 10 \\
            Faithfulness $\beta$ & 1000000 \\        
            Importance Minimality $\beta$ & 0.05 \\
            $P_{start}$ & 0.9 \\
            $P_{final}$ & 0.1 \\
            Anneal P? & True \\
            Last Token Only? & False \\
            \bottomrule\\
        \end{tabular}
        \caption{Hyperparameters for the GPT2 decomposition task}
        \label{tab:gpt2-decomp-hparams}
\end{figure}
\subsubsection{Fact Location}
\begin{figure}[h!]
    \centering
    \begin{minipage}{0.48\linewidth}
        \centering
        \includegraphics[width=\linewidth]{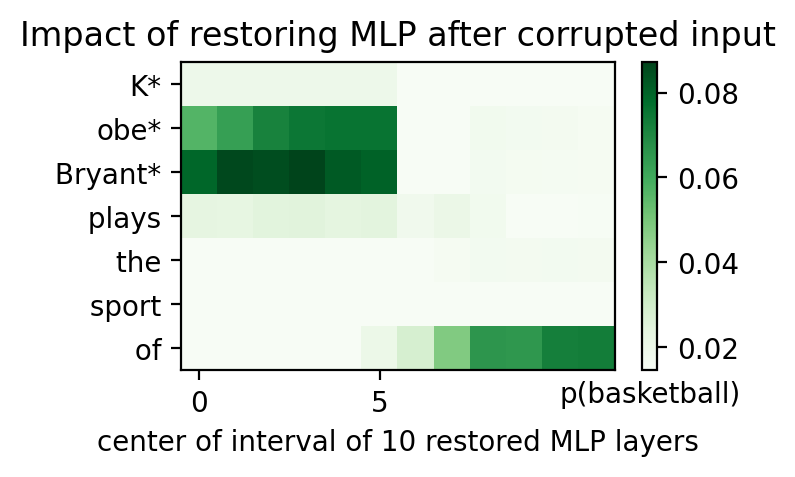}
        \caption{Causal tracing on the Kobe Bryant example}
        \label{fig:kobe}
    \end{minipage}
    \hfill
    \begin{minipage}{0.48\linewidth}
        \centering
        \includegraphics[width=\linewidth]{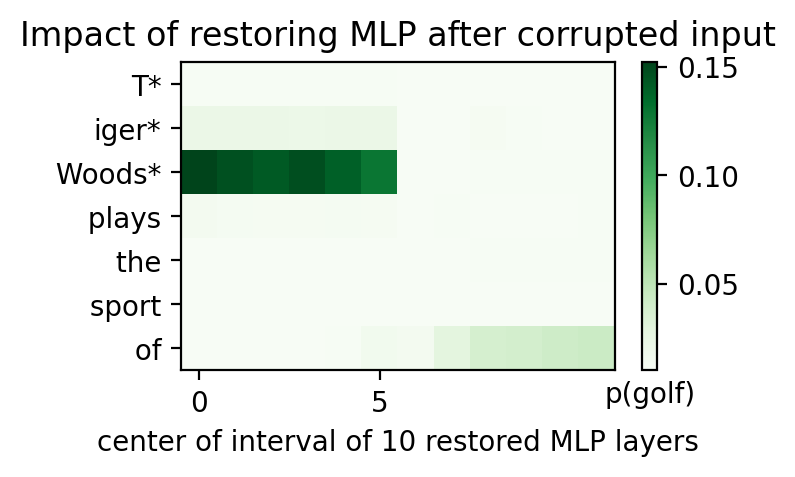}
        \caption{Causal tracing on the Tiger Woods example}
        \label{fig:tiger}
    \end{minipage}
\end{figure}

\begin{figure}[h!]
    \centering
    \includegraphics[width=0.5\linewidth]{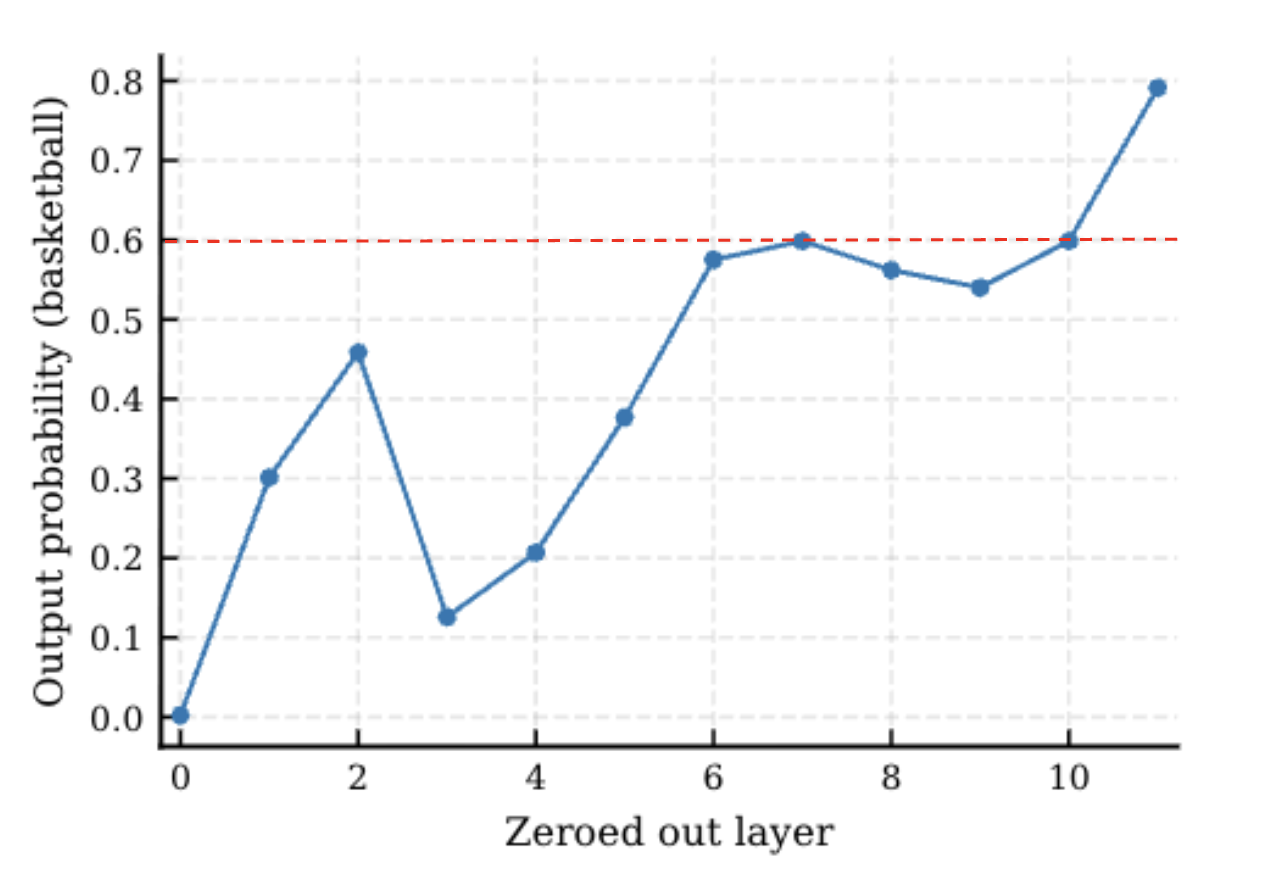}
    \caption{The effect of zeroing out entire MLP-layers in GPT2-small on the "Kobe Bryant plays the sport of" example. Dashed line is the full model's output probability.}
    \label{fig:zeroing_gpt2}
\end{figure}

 Figure \ref{fig:kobe} and \ref{fig:tiger} show the result of running causal tracing from Meng et al.\cite{meng2023locatingeditingfactualassociations} on our 2 fact-examples. Causal tracing suggests that the fact can be edited all the way from layer 0 MLPs to the layer 5 MLPs. In our findings, we see evidence that the athlete $\rightarrow$ sport association is present as early as after the first MLP layer. The fact is editable up until the fifth MLP layer, after which the information is likely moved across the sequence with attention.

Editing a fact comes down to correlating a specific read direction $\vec{U}$ ($\vec{K}$ in \cite{meng2023locatingeditingfactualassociations}) with a desired write direction $\vec{V}$. Under that framing, it is easy to see how \textit{any} MLP that encounters a certain fact before the unembedding. In Meng et al.\cite{meng2023locatingeditingfactualassociations} they edit facts by solving a small optimisation problem to yield $\vec{V}$ that maximises the probability of the fact they want to predict. The key-vector $\vec{K}$ -- the input to the MLP -- could in theory be anything. In later layers the \textit{basketball}-concept may have already been written to the residual stream and we are introducing a mapping from basketball to e.g. football.  

In figure \ref{fig:zeroing_gpt2} we take this to the extreme, and show that you can actually entirely zero-out the layer in which it is most beneficial to edit the fact, showing conclusively that the fact does not live there. Note that the drop in probability to 0 when zeroing out layer 0 does \textit{not} imply the fact lives there -- zeroing out this layer simply destroys the model.
\subsubsection{Partial losses}
\label{sec:cheating}
When decomposing large models over very few samples, the decomposed model is likely to be much smaller than the original model. This means that the decomposed model has room to cheat and simply learn a function that produces the same output, by creating new mechanisms and putting them in the components that are unused in the original model. It can then drive all other causal importances to 0 through optimising for minimality. This creates a decomposition that is likely more minimal, but does not teach us anything about the original model. Specifically, we often see this in the form of the model maintaining one layer (the last) with a suboptimal amount of subcomponents active, such that it can keep the other layers very sparse. If you consider a model decomposed over \textit{just} the example "Tiger Woods plays golf", it is easy to see how the model could just learn a handful of rank-1 subcomponents in the final layer, that writes the relevant token directions onto the residual stream, instead of learning the underlying mechanism. When decomposing over larger datasets, this effect is minimised as the cheating subcomponents would incur heavy reconstruction losses for all other examples.

We circumvent this with the 2 losses from section 6.2:

\begin{align}
\mathcal{L}_{recon\_partial}
= D_{KL}\!\left(
f\!\big(x \mid W^{1},\ldots,W^{l \in \mathcal{M}}(x,g^{l}(x)),\ldots,W^{L}\big),\;
f(x \mid W)
\right),
\end{align}

\begin{align}
\mathcal{L}_{stochastic\_recon\_partial}
= \frac{1}{S} \sum_{s=1}^S 
D_{KL}\!\left(
f\!\big(x \mid W^{1},\ldots,W^{l \in \mathcal{M}^{(s)}}(x,r^{l,(s)}),\ldots,W^{L}\big),\;
f(x \mid W)
\right),
\end{align}

where $\mathcal{M}\subseteq \{1,...,L\}$.

These are equivalent to $\mathcal{L}_{recon}$ and $\mathcal{L}_{stochastic\_recon}$, but they sample only some subset of layers in the model to replace and leave everything else in the target model intact. This severely reduces the ability for the model to implement cheating mechanisms in any layer, as it \textbf{1)} may not be able to rely on that layer and \textbf{2}) these mechanisms might clash with the weight matrices of the full model.

\newpage
\subsubsection{Fact dataset}
\label{sec:facts}
\begin{figure}[h!]
\begin{lstlisting}
Serena Williams is a professional tennis player
The Great Wall of China is in China
The Mona Lisa was painted by Leonardo da Vinci
Mount Everest is the tallest mountain on Earth
The Eiffel Tower is located in Paris France
Usain Bolt is the fastest man in the world
The Pacific Ocean is the largest ocean on Earth
Albert Einstein developed the theory of relativity
J.K. Rowling wrote the Harry Potter series
The Amazon River flows through South America
William Shakespeare wrote Romeo and Juliet
Beethoven composed the Fifth Symphony
The Sahara Desert is in Africa
Apple Inc. was founded by Steve Jobs Steve Wozniak and Ronald Wayne
The sun is a star
The Beatles were a British rock band
Cristiano Ronaldo is a professional soccer player
The Statue of Liberty is in New City
Pablo Picasso was a famous painter
The Nile is the longest river in the world
Charles Darwin developed the theory of evolution
Google was founded in 1998
The Taj Mahal is in India
The Titanic sank in 1912
Jane Austen wrote Pride and Prejudice
The Moon orbits the Earth
Mount Kilimanjaro is in Tanzania
The Leaning Tower of Pisa is in Italy
Walt Disney created Mickey Mouse
Vincent van Gogh painted Starry Night
\end{lstlisting}
\caption{Excerpt from GPT-5 generated "fact" dataset.}
\end{figure}

\newpage
\subsection{Training routine}
\label{sec:step}
\begin{algorithm}[H]
\caption{SPD Training Step (Single Minibatch)}
\label{alg:spd-train}
\begin{algorithmic}[1]
\Require Original model $\mathcal{M}_\theta$
\Require Decomposed model $\tilde{\mathcal{M}}_\phi$
\Require Causal-importance routine $\mathcal{C_{\phi}}$
\Require Dataset $\mathcal{D}$
\Require Loss functions $\{\mathcal{L}_k\}_{k=1}^n$ with coefficients $\{\lambda_k\}_{k=1}^n$

\State Sample minibatch $X \subset \mathcal{D}$
\State $Y \gets \mathcal{M}_\theta(X)$ ~~~~~// run original model
\State $C \gets \mathcal{C_{\phi}}(X)$ ~~~~~// compute causal importances on this batch
\State $R \gets \text{SampleMasks}(C)$ ~~~~~// sample stochastic subcomponent masks
\State $\tilde{Y} \gets \tilde{\mathcal{M}}_\phi(X; R)$ ~~~~~// run decomposed model with masks
\State $L \gets 0$
\For{$k = 1 \dots n$}
    \State $L \gets L + \lambda_k \, \mathcal{L}_k(X, Y, \tilde{Y}, C, R, \mathcal{M}_\theta, \tilde{\mathcal{M}}_\phi$)
\EndFor
\State Update parameters of $\tilde{\mathcal{M}}_\phi$ and $\mathcal{C}_{\phi}$
\end{algorithmic}
\end{algorithm}

\end{document}